%% file: lipics-v2021-sample-article.tex
\documentclass[a4paper,UKenglish,cleveref, autoref, thm-restate]{lipics-v2021}

\usepackage{times}
\usepackage{soul}
\usepackage{url}
\usepackage[utf8]{inputenc}
\usepackage{graphicx}
\usepackage{amsmath}
\usepackage{amsthm}
\usepackage{booktabs}
\usepackage{algorithm}
\usepackage{algorithmic}
\usepackage{subcaption}
\usepackage{caption}
\usepackage{float}
\usepackage[export]{adjustbox}
\usepackage{threeparttable}
\usepackage{multirow}
\usepackage{multicol}
\usepackage{xcolor}
\usepackage{newfloat}
\usepackage{listings}
\usepackage{amssymb}

\title{ParLS-PBO: A Parallel Local Search Solver for Pseudo Boolean Optimization} 

\author{Zhihan Chen}{Key Laboratory of System Software (Chinese Academy of Sciences) and State Key Laboratory of Computer Science, Institute of Software, Chinese Academy of Sciences, Beijing, China \and School of Computer Science and Technology, University of Chinese Academy of Sciences, Beijing, China}{chenzh@ios.ac.cn}{https://orcid.org/0000-0001-5702-2508}{}

\author{Peng Lin}{Key Laboratory of System Software (Chinese Academy of Sciences) and State Key Laboratory of Computer Science, Institute of Software, Chinese Academy of Sciences, Beijing, China \and School of Computer Science and Technology, University of Chinese Academy of Sciences, Beijing, China}{linpeng@ios.ac.cn}{https://orcid.org/0009-0002-4183-5998}{}

\author{Hao Hu}{Key Laboratory of System Software (Chinese Academy of Sciences) and State Key Laboratory of Computer Science, Institute of Software, Chinese Academy of Sciences, Beijing, China}{huhao@ios.ac.cn}{}{}

\author{Shaowei Cai\footnote{Corresponding author}}{Key Laboratory of System Software (Chinese Academy of Sciences) and State Key Laboratory of Computer Science, Institute of Software, Chinese Academy of Sciences, Beijing, China \and School of Computer Science and Technology, University of Chinese Academy of Sciences, Beijing, China \and \url{https://lcs.ios.ac.cn/~caisw/}}{caisw@ios.ac.cn}{https://orcid.org/0000-0003-1730-6922}{}

\authorrunning{Z. Chen, P. Lin, H. Hu and S. Cai} 

\Copyright{Zhihan Chen, Peng Lin, Hao Hu and Shaowei Cai} 



\category{} 

\relatedversion{} 

\nolinenumbers 
\funding{This work is supported by National Key R\&D Program of China (2023YFA1009500).}
\supplement{}
\supplementdetails[subcategory={}, cite={}, swhid={}]{Software}{https://github.com/shaowei-cai-group/ParLS-PBO.git}

\EventEditors{Paul Shaw}
\EventNoEds{1}
\EventLongTitle{30th International Conference on Principles and Practice of Constraint Programming (CP 2024)}
\EventShortTitle{CP 2024}
\EventAcronym{CP}
\EventYear{2024}
\EventDate{September 2--6, 2024}
\EventLocation{Girona, Spain}
\EventLogo{}
\SeriesVolume{307}
\ArticleNo{8}

\begin{document}

\input{0_macros}

\maketitle

\input{0_Abstract}

\input{1_Introduction}

\input{2_Preliminary}

\input{2.5_Improving_LSPBO}
\input{3_Parelle_Methods}


\input{5_Experimental_Results}

\input{6_Conclusion}

\bibliography{ref}

\end{document}

%% file: 0_macros.tex

\def\hu#1{\textcolor{blue}{#1}}
\def\cai#1{\textcolor{red}{#1}}
\def\chen#1{\textcolor{orange}{#1}}

\def\lspbo{\textit{LS-PBO}}
\def\nupbo{\textit{NuPBO}}
\def\lspbodeci{\textit{DeciLS-PBO}}
\def\lspbodym{\textit{DLS-PBO}}
\def\hybrid{\textit{HYBRID}}
\def\pboihs{\textit{PBO-IHS}}
\def\gurobi{\textit{Gurobi}}
\def\parlspbo{\textit{ParLS-PBO}}
\def\fiberscip{\textit{FiberSCIP}}
\def\scip{\textit{SCIP}}
\def\cplex{\textit{CPLEX}}
\def\parascip{\textit{ParaSCIP}}
\def\parLSPBO{\textit{ParLS-PBO}}

\def\score#1{score(#1)}
\def\hscore#1{hscore(#1)}
\def\oscore#1{oscore(#1)}
\def\scorenew#1{score^{*}(#1)}

\def\ratingmixed#1{r_{mix}(#1)}

\def\aboolvar#1{x_{#1}}
\def\aliteral#1{l_{#1}}

\def\acoeff#1{a_{#1}}
\def\adegree#1{b_{#1}}
\def\acostcoeff#1{c_{#1}}

\def\asolution{\alpha}
\def\funobj#1{obj(#1)}
\def\valueobjbest{obj^{*}}

\def\aweight#1{w(#1)}
\def\ahardcon{hc}
\def\objcon{oc}

\def\rounds{K}

\def\penaltyweight{p}

%% file: 0_Abstract.tex
\begin{abstract}
As a broadly applied technique in numerous optimization problems, recently, local search has been employed to solve Pseudo-Boolean Optimization (PBO) problem.
A representative local search solver for PBO is~\lspbo{}. In this paper, firstly, we improve ~\lspbo{} by a dynamic scoring mechanism, which dynamically strikes a balance between score on hard constraints and score on the objective function.

Moreover, on top of this improved ~\lspbo{}, we develop the first parallel local search PBO solver. 
The main idea is to share good solutions among different threads to guide the search, by maintaining a pool of feasible solutions. For evaluating solutions when updating the pool, we propose a  function that considers both the solution quality and the diversity of the pool. Furthermore, we calculate the polarity density in the pool to enhance the scoring function of local search.  
Our empirical experiments show clear benefits of the proposed parallel approach, making it competitive with the parallel version of the famous commercial solver \gurobi{}.
\end{abstract}

%% file: 1_Introduction.tex
\section{
{Introduction}}


With the recent impressive progress in high-performance Boolean Satisfiability (SAT) and Maximum Boolean Satisfiability (MaxSAT) solvers,
increasing real-world problems are solved with the Conjunctive Normal Form (CNF) encoding.
However, in practice, 
CNF is ineffective in dealing with cardinality constraints, 
resulting in its size growing dramatically~\cite{STACS94-Benhamou-ProofCardinality}.
As a rich subject in various fields, Pseudo-Boolean Optimization (PBO) provides a better formalization than CNF in expressive power,
with the use of Linear Pseudo-Boolean (LPB) constraints.
Meanwhile, LPB constraints stay close to CNF and can benefit from advances in SAT solving~\cite{Chapter21-Roussel-PBandCardinality}.
The PBO problem is to find an assignment satisfying all LPB constraints that maximizes the objective function given.


%
Briefly, there are three categories of complete algorithms to solve the PBO problem. The first one is the linear search, which extends the PB solver by adding a constraint enforcing to find a better solution (in terms of the objective function value) when finding a solution satisfying all constraints~\cite{TR95-Barth-DPLinearPBO}. Several well-known PBO solvers are based on this idea, 
including \textit{Sat4j}~\cite{le2010sat4j}, \textit{RoundingSAT}~\cite{IJCAI18-Elffers-RoundingSAT}, and \hybrid{}~\cite{AAAI21-Devriendt-HybridPBO}. The second one is Branch-and-Bound, which focuses on the techniques to estimate the lower bounds of the objective value, as the search can be pruned whenever the lower bound is greater than or equal to the upper bound. The symbolic techniques to determine the lower bounds include Maximum Hitting Set~\cite{DAC95-Coudert-BnBPBOMHS} and Linear Programming Relaxation~\cite{DAC97-Liao-BnBPBOLPR}. The third one is to call the SAT solvers after encoding PB constraints into the CNF formula, such as \textit{MINISAT+}~\cite{een2006translating} and \textit{OpenWBO}\textit~\cite{SAT14-Martins-OpenWBO}. Moreover, mixed-integer programming (MIP) solvers can be directly applied to solve the PBO problem, as PB constraints can be treated as 0-1 linear constraints, representative solvers include \scip{}~\cite{bestuzheva2021scip} and \gurobi{}~\cite{gurobi2021gurobi}.


Complete algorithms often suffer from the scalability issue, 
which motivates the development of incomplete algorithms.
A typical incomplete approach is local search, which has been successfully used in many problems, 
including SAT{~\cite{DBLP:conf/sat/LiL12,DBLP:conf/sat/BalintS12,DBLP:conf/sat/CaiLS15}}, 
MaxSAT{~\cite{DBLP:conf/ijcai/LeiC18,DBLP:conf/aaai/Chu0L23}}, etc.  
Nevertheless, the literature on local search algorithms to solve PBO problem is quite limited.
The first local search-based PBO solver was proposed in~\cite{SAT21-Lei-LSPBO}, called ~\lspbo{}.  This local search solver introduced a constraint weighting scheme and a scoring function considering both hard and soft constraints to select Boolean variables to flip.
Later, \lspbo{} was improved by using a unit propagation-based method to produce better initial assignments~\cite{DBLP:journals/corr/abs-2301-12251}, resulting in the \lspbodeci{} solver. Very recently, on top of ~\lspbo{}, Chu et. al. developed \nupbo~\cite{chu2023towards}, which established the latest state-of-the-art local search based PBO solving. 
Additionally, Iser et. al. proposed an oracle-based local search approach in the context of PBO~\cite{DBLP:conf/ecai/IserBJ23} , which outperforms on various benchmark domains clearly the recent pure stochastic local search approach.

Recently, with the evolution of multi-core processors, 
parallel solving received growing interest.
The SAT competition\footnote{\url{http://www.satcompetition.org/}} set up a parallel track from 2009, 
while Satisfiability Modulo Theories (SMT) competition\footnote{\url{https://smt-comp.github.io/}} introduces parallel tracks in 2021. 
{In short,} parallel algorithms contain two major directions.
The first one is based on the concept of divide-and-conquer, which divides the problem into several 
sub-problems, 
and each thread solves sub-problems. 
For example, Treengeling~\cite{fleury2020cadical} is a representative SAT solver of this kind.
Meanwhile, commercial solvers, such as \textit{CPLEX}\footnote{\url{http://www.cplex.com/}} and \gurobi{}~\cite{gurobi2021gurobi} 
also implement their parallel versions via this approach. 
The other parallel approach is to integrate different solvers, including a solver with different configurations, and each thread runs a solver.
This approach is commonly known as portfolio, which is simple but effective.
The portfolio-based parallel SAT solvers, such as \textit{PRS}~\cite{chen2023prs}, \textit{P-mcomsps}~\cite{DBLP:conf/sat/FriouxBSK17}, and  \textit{Pakis}~\cite{tchinda2021hkis}, 
dominate the parallel track of SAT competitions in recent years. The parallel MaxSAT solvers~\cite{martins2011exploiting, martins2012parallel} based on the portfolio method also demonstrate a strong ability to efficiently solve a large number of problem instances due to the use of complementary search strategies and sharing learned clauses between threads.



{In this paper, at first,
we improve the typical~\lspbo{} solver by introducing a new dynamic scoring mechanism,
which can somehow avoid the local optimum situation even after flipping thousands of variables. This leads to an improved algorithm called \lspbodym{}.
Then, based on the \lspbodym{}, we develop~\parlspbo{}, to the best of our knowledge, 
the first parallel local search-based PBO solver. 
Our parallel solver runs different local search procedures in the worker threads and maintains a solution pool, 
which collects good feasible solutions from the working threads and in turn can be used to guide the local search procedures.
During the search process, the solution pool is updated by adding new solutions and removing solutions from it.  
To update the solution pool, we propose a function to measure the feasible solutions found by local search, which considers both the objective value of the solution and the diversity of the solutions in the pool.

The most important part of our parallel solver is how to use the solution pool to help the local search. 
In this work, this is done in two ways. 
Firstly, the solutions in the pool can be directly used to help local search when it stagnates for a long time. 
Specifically, in such a situation, a local search process of a thread restarts from a good feasible solution picked from the pool. Secondly, we calculate the polarity density (the proportions of being 1 and 0) for each variable 
{once a solution is added into the solution pool.} 
This polarity density information is used to enhance the scoring function of local search when picking the variable to flip in each step.
The intuition is when a certain polarity (either 1 or 0) of a variable occurs in most high-quality solutions, 
it brings preference to assign the variable to that polarity.


We carry out experiments to evaluate our algorithms~\lspbodym{} and~\parlspbo{} on both real-world applications encoded benchmark and standard benchmarks, compared with state-of-the-art solvers including \lspbo{}~\cite{SAT21-Lei-LSPBO}, \lspbodeci{}~\cite{DBLP:journals/corr/abs-2301-12251}, \nupbo{}~\cite{chu2023towards},  \scip{}~\cite{bestuzheva2021scip}, ~\hybrid{}~\cite{AAAI21-Devriendt-HybridPBO}, \pboihs{}~\cite{DBLP:conf/sat/0003BJ22}, \gurobi{}~\cite{gurobi2021gurobi}), and \fiberscip{}~\cite{DBLP:journals/informs/ShinanoHVW18}. 
Our results show that our parallel solver has significantly better performance than all sequential solvers, and competes well with the parallel versions of \gurobi{}.
Furthermore, \parlspbo{} shows good scalability up to 32 threads as its performance improves with the number of threads.


The remainder of this paper is structured as follows. 
Section 2 introduces preliminary knowledge. 
Section 3 analyzes the weakness of \lspbo{} and introduces an improved solver \lspbodym{}. Section 4 presents the proposed parallel solver \parlspbo{}. Experimental studies are presented in Section 5. 
Finally, we give some concluding remarks in Section 6.


%% file: 2_Preliminary.tex
\section{Preliminaries}  

\subsection{Pseudo-Boolean Optimization}


A Boolean variable $\aboolvar{i}$ can take only two values \textit{false} and \textit{true}, or equivalently $\{0, 1\}$.
A literal $\aliteral{i}$ is either a variable $\aboolvar{i}$ or its negation $\neg \aboolvar{i}$.
Given a set of $n$ Boolean variables $\{\aboolvar{1}, \dots, \aboolvar{n}\}$, 
a \textit{linear pseudo-Boolean constraint} (LPB constraint) is formed as follows:
\begin{equation*}
    \sum_{i=1}^{n} \acoeff{i} \cdot \aliteral{i}  \triangleright \adegree{}, \quad \acoeff{i}, \adegree{} \in \mathbb{Z}, \quad \triangleright \in \{=,\leq, <, \geq, >\}
\end{equation*}
where $\acoeff{i}$ is the coefficient for literal $\aliteral{i}$, $\adegree{}$ is called as the degree of the constraint, and $\triangleright$ is one of the classical relational operators.
With a given assignment or partial assignment, the constraint is satisfied if its left and right terms satisfy the relational operator.
Otherwise, it is unsatisfied.

Moreover, replacing all literals $\aboolvar{i}$ (respectively $\neg \aboolvar{i}$) with negative coefficients with $1 - \neg \aboolvar{i}$ (respectively $1-\aboolvar{i}$), 
a LPB constraint can be normalized into the following form~\cite{Chapter21-Roussel-PBandCardinality}:
\begin{equation*}
    \sum_{i=1}^{n}\acoeff{i} \cdot \aliteral{i} \geq \adegree{}, \quad \acoeff{i}, \adegree{} \in \mathbb{N}_{0}^{+}
\end{equation*}

Given a conjunction of LPB constraints, \textit{Pseudo-Boolean Solving} (PBS) problem is a decision problem to find an assignment such that all constraints are satisfied.
\textit{Pseudo-Boolean Optimization} (PBO) problem is an optimized version of the PBS problem, 
aiming to find an assignment satisfying all constraints with the minimal value of a given objective function.
In this paper, we focus on the PBO problem consisting of a conjunction of LPB constraints and a linear objective function.
Therefore, a PBO instance subjecting to $m$ LPB constraints has the following form:
    \begin{alignat*}{2}
        &\!\min_{\{\aboolvar{1},\dots,\aboolvar{n}\}} &\quad \sum_{i=1}^{n} \acostcoeff{i} \cdot \aliteral{i}, &\quad \acostcoeff{i} \in \mathbb{Z} \\
        &\text{subject to:} &\quad \bigwedge_{j=1}^{m} \sum_{i=1}^{n}\acoeff{ji} \cdot \aliteral{i} \geq \adegree{j}, &\quad \acoeff{ji}, \adegree{j} \in \mathbb{N}_{0}^{+}
    \end{alignat*}
where $\acostcoeff{i}$ is the objective coefficient for literal $\aliteral{i}$.









\subsection{A Review of \lspbo{} Solver}

\lspbo{} is a representative local search solver for PBO, and serves as the basis of other local search PBO solvers.
Briefly, it contains two main ideas: a  \textit{Constraint Weighting Scheme} and  \textit{Scoring Functions} for guiding the search process.

To solve a standard PBO instance, 
\lspbo{} proposed a soft \textit{objective constraint}: $\sum_{i=1}^{n}\acostcoeff{i} 
\cdot \aliteral{i} < \valueobjbest{}$, where $\valueobjbest{}$ indicates the objective value of the best solution in the current run,
and other constraints are set as hard.
\lspbo{} uses a weighting technique to increase the weights of falsified constraints,
so that the search process is biased toward satisfying them.
Specifically, it used dynamic weights (denoted as $\aweight{\cdot}$) to help the search avoid stuck in the local optimum, 
while increasing the weights of hard constraints to find feasible solutions, 
and the weight of objective constraint to find better solutions.

Besides, scoring functions are essential in local search algorithms to guide the search process,
which typically measures the benefits of flipping a Boolean variable.
In \lspbo{}, the score of flipping a variable $\aboolvar{}$ (denoted as $\score{\aboolvar{}}$) was defined as follow:
\begin{equation}
    \centering
    \label{eq:2_2_flip_score_fun}
    \score{\aboolvar{}} = \hscore{\aboolvar{}} + \oscore{\aboolvar{}}
\end{equation}
where $\hscore{\aboolvar{}}$ indicates the decrease of the total penalty of falsified hard constraints caused by flipping $\aboolvar{}$,
and $\oscore{\aboolvar{}}$ indicates the decrease of the penalty of the objective constraint caused by flipping $\aboolvar{}$.
In detail, the penalty of falsifying a hard constraint $\ahardcon{}$ was defined as $\aweight{\ahardcon{}} \cdot \max{(0, \adegree{} - \sum_{i=1}^{n}\acoeff{i} \cdot \aliteral{i})}$, 
and the penalty for the objective constraint $\objcon{}$ was defined as $\aweight{\objcon{}} \cdot \sum_{i=1}^{n}\acostcoeff{i} \cdot \aliteral{i}$.

%% file: 2.5_Improving_LSPBO.tex
\section{Improving \lspbo{} Solver with Dynamic Scoring Mechanism}






As introduced in the preliminary, the score of a filliping variable $\aboolvar{}$ ($\score{\aboolvar{}}$) in \lspbo{} is presented as Equation~\ref{eq:2_2_flip_score_fun}.
The algorithm selects the variable with the highest positive score,
indicating the biggest decrease in the penalty of hard constraints and objective constraint.
A drawback of \lspbo{} is the lack of dynamic adjustments to the ratio of the soft and hard constraints. If a feasible solution cannot be found within a certain period of time, the search mechanism should adaptively prioritize finding feasible solutions, thereby increasing the ratio attributed to the hard constraints. Conversely, if feasible solutions have been frequently found recently, then it would be beneficial to increase the ratio of the soft constraints to guide the search towards discovering better solutions.

To resolve this drawback, we introduce a new \textit{dynamic scoring function}, denoted as $\scorenew{\aboolvar{}}$, to adjust the significance of $\oscore{\aboolvar{}}$ 
for every given $\rounds{}$ steps ($\rounds{}$ is a parameter),  which is defined as follows:
\begin{equation}
    \scorenew{\aboolvar{}} = \hscore{\aboolvar{}} + \penaltyweight{} \cdot \oscore{\aboolvar{}}
\end{equation}
where $\penaltyweight{}$ is a dynamic ratio initially set as $1$.
It would be decreased as $\penaltyweight/inc$ (where $inc > 1$) if no feasible solution is found during the recent  $\rounds{}$ steps, to guide the search towards a feasible solution. 
Otherwise, respectively, it would be increased as $\penaltyweight \cdot inc$ when a feasible solution is found within the recent $\rounds{}$ steps, to guide the search process for a better solution.


\begin{example}
\label{example:2_5_dynamic_scoring}
Considering a PBO instance:  
    \begin{alignat*}{2}
        &\!\min_{\{\aboolvar{1},\aboolvar{2}, \aboolvar{3}\}} &\quad 10 \cdot \aboolvar{1} + 20 \cdot \aboolvar{2} + 30 \cdot \aboolvar{3}\\
        &\text{subject to:} &\quad 2 \cdot \aboolvar{1} + 3 \cdot \aboolvar{2} + 4 \cdot \aboolvar{3} \geq 5
    \end{alignat*}
and suppose current weights $\aweight{\ahardcon{}}$ and $\aweight{\objcon{}}$ are $2$ and $1$.
For the given assignment $(\aboolvar{1} = 1, \aboolvar{2} = 0, \aboolvar{3} = 0)$, 
the corresponding $\hscore{\cdot}$ and $\oscore{\cdot}$ are as follows:

\begin{table}[H]
\centering
\renewcommand\arraystretch{1.2}
\setlength\tabcolsep{12pt}
\begin{tabular}{c|ccc}
\hline
$\cdot$ & $\aboolvar{1}$ & $\aboolvar{2}$ & $\aboolvar{3}$ \\\hline
$\hscore{\cdot}$ & -4 & 6 & 6\\
 $\oscore{\cdot}$ & 10 & -20 & -30 \\
\hline
\end{tabular}
\end{table}

Consider the following two situations:
\begin{itemize}
    \item If feasible solutions are found frequently in recent period, the value of $p$ will gradually increase, guiding the search to lower the cost of the objective constraint. Suppose the current value of $p$ is 2, then $score^*(x_1) = 16, score^*(x_2)=-34, score^*(x_3)=-54$. In this case, $x_1$ will be picked and flipped, resulting in a decrease of 10 in the cost of the objective constraint. (even if it is not a feasible solution.)

    \item If the algorithm has not visited feasible solutions for a period ($\rounds{}$ steps), the value of $p$ will gradually decrease, guiding the search to find feasible solutions. Suppose the current value of $p$ is 0.1. then $score^*(x_1) = -3, score^*(x_2)=4, score^*(x_3)=3$. In this case, $x_2$ will be picked and flipped, resulting in a feasible solution.

\end{itemize}

\end{example}


We denote the improved version of~\lspbo{} solver with dynamic scoring mechanism as~\lspbodym{}.

%% file: 3_Parelle_Methods.tex
\begin{figure}[t]
    \centering
    \includegraphics[width=0.65\textwidth,height=0.5\textheight]{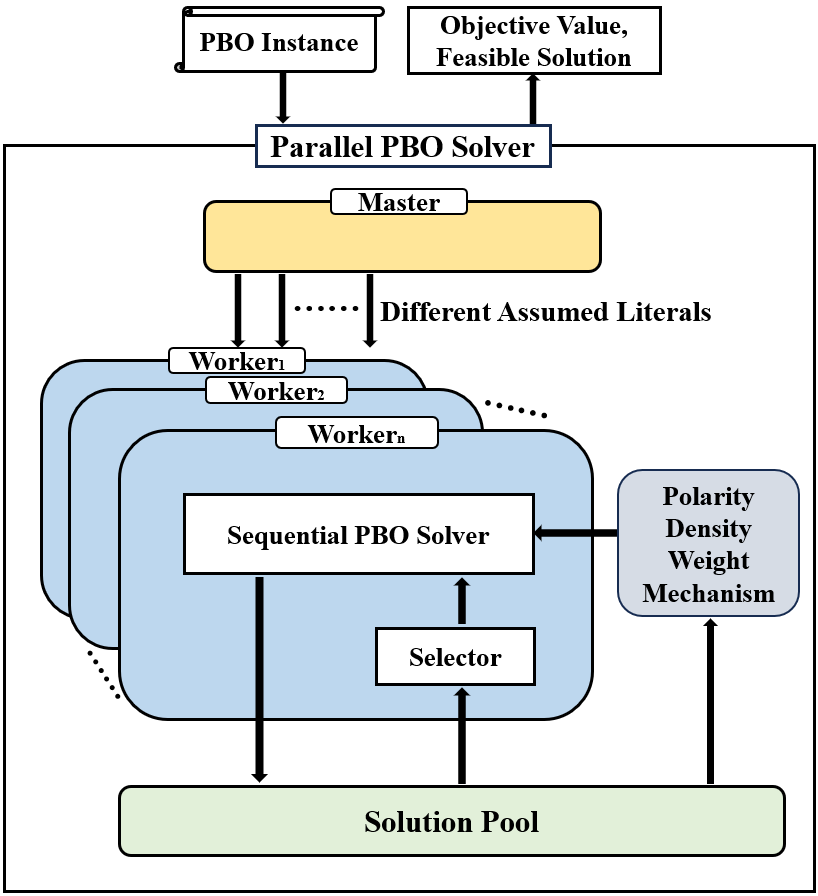}
    \caption{Architecture of the \parlspbo{}.}
    \label{fig:3_parallel_PBO_framework}
\end{figure}

\section{
{Parallel Local Search Solver for Pseudo-Boolean Optimization}}


In this section, we propose a parallel local search solver ~\parlspbo{}. 
The architecture of~\parlspbo{} is shown in Figure~\ref{fig:3_parallel_PBO_framework}, 
which consists of two major contributions: \textit{Solution Pool} and \textit{Polarity Density Weight}.
We first describe the global framework of~\parlspbo{}, 
then we present the contributions in detail separately.


\subsection{Framework 
{of~\parlspbo{} Solver}}

As a portfolio-based local search PBO solver,~\parlspbo{} contains a \textit{master} thread and multiple \textit{worker} threads.
The master thread reads the input PBO instance,
and then produces different initial partial assignments via \textit{literal assume} technique for worker threads; finally, when the time limit is reached, it outputs the best solution returned from all worker threads.
In detail, supposing there are $T$ worker threads, the master thread selects $\lceil \frac{T}{2} \rceil$ random variables. 
Then, for each variable selected $\aboolvar{i}$, it generates a positive literal $\aboolvar{i}$ and a negative literal $\neg\aboolvar{i}$. Therefore, it generates $T$ (or $T+1$ if $T$ is odd) different assumed literals in total for worker threads.


Each worker thread receives an assumed literal $\ell$ (either $x_i$ or $\neg x_i$) and applies the \textit{unit propagation}~\cite{marques2021conflict} technique to simplify the formula. Note that a solution found by local search for such a formula can be directly transformed to a solution for the original PBO instance, by adding the assumed literal as the value for the corresponding variable, and adding the value for reduced variables via unit propagation if any. 
Then the worker thread launches a local search solver to solve the PBO instance.
In default, the worker thread launches~\lspbodym{}.
To bridge different worker threads, we propose a \textit{Solution Pool} to share high-quality feasible solutions found from different worker threads. 
When the search process of a worker thread is blocked in the local optimum after flipping a certain number of variables,
it attempts to restart with a high-quality feasible solution in the solution pool.
Furthermore, we introduce the concept of \textit{Polarity Density Weight} with the intuition of preference of certain polarity of a variable if it occurs in most high-quality solutions.




\subsection{
{Maintaining the Solution Pool}}

The solution pool aims to collect good feasible solutions, preferring those with more differences. 
To this end, we consider a mixed quality rating function $\ratingmixed{\cdot}$ by measuring two terms together:
the quality in objective value, and the diversity.
For a feasible solution $S$, the rating function $\ratingmixed{S}$ w.r.t. a solution pool is defined as follows: 
\begin{equation}
    \ratingmixed{S} = rank_{obj}(S) \cdot p^{*} + rank_{div}(S) \cdot (1-p^{*})
\end{equation}
where $rank_{obj}(\cdot)$ and $rank_{div}(\cdot)$ represent the ranking of $S$ in the solution pool in objective value and diversity value. Specifically, for the objective value, the solution with the minimal objective value is considered as the best solution, hence its $rank_{obj}(\cdot)$ value is 1. While for the diversity value, the solution with the maximum diversity value is considered the best, thus its $rank_{div}(\cdot)$ value is 1. $p^{*}$ is a penalty parameter within $[0, 1]$ to adjust the significance of the objective value term and diversity term.

The difference between two solutions is measured as the sum of the number of different polarities,
and the diversity value of a solution $S$ w.r.t. The solution pool is measured as the sum of differences between $S$ and all other solutions in the solution pool.  Formally,
$$div(S)=\sum_{S'\in \mathcal{P}} Hamming(S,S')$$

When a worker thread finds a new feasible solution $S$, 
if the solution pool is not full, then just add it.
Otherwise, $S$ replaces the worst one (the solution with the biggest $r_{mix}(\cdot)$ value).

We note that the $\ratingmixed{S}$ function in this work resembles a previous population management strategy ~\cite{Chen-2016-rank}. We focus on the ranking rather than the value, which can be seen as a normalization.


        



\subsection{
{Using the Solution Pool to Guide the Search}}
In this subsection, we discussed how the solution pool guides each worker, including replacing solutions with better solutions from the solution pool when a worker being trapped, and utilizing the variable polarity preference in the solution pool to influence the selection of variables to flip during the search process.
\subsubsection{Solution Sharing Strategy}

When a worker thread fails to find a better feasible solution for a while, which means that it may be trapped in a local optimum, it selects a feasible solution with a smaller objective value from the solution pool and replaces the current one. 

In practice, each worker thread preserves the current best feasible solution (denoted as $S^{*}$) as well as the corresponding objective value $\valueobjbest{}$.
When the search process fails to find a better solution after $R$ steps, it picks a solution from the solution pool as a new starting point.  

To prevent excessive overlap of search spaces among various threads, we employ a probability-based method to select solutions in the pool, rather than directly choosing the best solution in the pool. Specifically, let $\{S_1,\dots,S_k\}$ denotes the set of feasible solutions in the solution pool with objective values not bigger than $obj^{*}$ (the set will not be empty, as it at least contains $S^{*}$),
and $\Delta_{i}$ denotes the difference between the objective value of $S_i$ and $\valueobjbest{}$.
Then the probability of selecting $S_i$ is $\Delta_{i} /\sum_{j=1}^{k}{\Delta_{j}}$.


    
        


\subsubsection{Polarity Density Weight}
Besides using the solutions in the pool to guide the search process directly when it gets stuck, we propose a deeper guiding method, which utilizes a piece of valuable hidden information in the solution pool --- the occurrence of polarities (0 or 1) of variables.
To measure the effect of this kind of information, we propose the concept of \textit{polarity density weight} for a variable $\aboolvar{}$, denoted as $w_{pd}(\aboolvar{})$,
which reflects the preference of certain polarity of $\aboolvar{}$ appearing in high-quality solutions.

In detail, for a variable $\aboolvar{}$, $w_{pd}(\aboolvar{})$ is initialized as $1$.
Once a high-quality feasible solution $S$ is added into the solution pool, $w_{pd}(\aboolvar{})$ will add (respectively, minus) a step value $\beta$ when positive (respectively,  negative) polarity of $\aboolvar{}$ appears in $S$.
In fact, via this updating mechanism, the higher (respectively, lower) value in $w_{pd}(\aboolvar{})$ indicates the higher preference of positive (respectively, negative) polarity for $\aboolvar{}$ in high-quality solutions.
To limit the influence of $w_{pd}(\aboolvar{})$ and avoid possible calculation problems in negative values, 
we restrict $w_{pd}(\aboolvar{})$ into an interval of $[1-\epsilon, 1+\epsilon]$, where $\epsilon{}$ scales the bound.
Therefore, the $w_{pd}(\aboolvar{})$ is updated as follows:
\begin{equation}
    \centering
    w_{pd}(x)=
    \begin{cases}
        \max(w_{pd}(x) - \beta, 1 - \epsilon), & \text{if }x=0 \text{ in } S\\
        \min(w_{pd}(x) + \beta, 1+\epsilon), & \text{if }x=1 \text{ in } S 
    \end{cases}
\end{equation}



The polarity density weight is used to enhance the scoring function of picking a variable to flip during the search process. The resulting enhanced scoring function, denoted as
 $score^{**}(\aboolvar{})$, is defined as follows:
\begin{equation}
    \centering
    score^{**}(\aboolvar{})\!=\! 
    \begin{cases}
        \scorenew{\aboolvar{}}\!\cdot\!w_{pd}(\aboolvar{}),&\text{if }\aboolvar{}\!=\!0\text{ in }S_{cur} \\
        \scorenew{\aboolvar{}}/w_{pd}(\aboolvar{}),&\text{if }\aboolvar{}\!=\!1\text{ in }S_{cur}
    \end{cases}
\end{equation}
where $S_{cur}$ is the current assignment maintained by the local search process.

The multiplication of polarity density weight influences the flip of a variable $\aboolvar{}$ from $0$ to $1$, 
as it increases the combined score if the preference of positive polarity exists ($w_{pd}(\aboolvar{}) > 1$) to guide the search process to realise the flip.
Respectively, in reverse, the division of polarity density weight influences the flip from $1$ to $0$.


\begin{example}
Continuing with Example 1, Suppose that most of the solutions that entered the solution pool have the assignment $(\aboolvar{1}=1, \aboolvar{2}=1, \aboolvar{3}=0)$, resulting in $w_{pd}(\aboolvar{1})=1.1, w_{pd}(\aboolvar{2})=1.1, w_{pd}(\aboolvar{3})=0.9$.

For the given assignment $(\aboolvar{1}=1, \aboolvar{2}=0, \aboolvar{3}=0)$ and $p=1$, the corresponding $score^*(\cdot)$ can be calculated as: $score^*(\aboolvar{1})=6,score^*(\aboolvar{2})=-14,score^*(\aboolvar{3})=-24$. 

Then $score^{**}(\aboolvar{1})=6 \div 1.1, score^{**}(\aboolvar{2})=(-14) \times 1.1, score^{**}(\aboolvar{3})=(-24) \times 0.9$.


\end{example}

%% file: 5_Experimental_Results.tex
\section{Experiments}
The experiments are organized as three  parts.
At first, we focus on comparing \lspbodym{},~\parlspbo{} with state-of-the-art solvers including commercial solvers. 
Secondly, we analyze the effectiveness of the strategies to guide the search via the solution pool in~\parlspbo{}. 
Finally, we present the tendency in performance of~\parlspbo{} with the increase of the number of threads.
Source code and detailed results are made publicly available on GitHub\footnote{\url{https://github.com/shaowei-cai-group/ParLS-PBO.git}}.

\subsection{Benchmark}

\begin{itemize}
    \item \textbf{Real-World}: Three real-world application problems, which are presented in the literature~\cite{SAT21-Lei-LSPBO}, including the Minimum-Width Confidence Band Problem~\cite{berg2017minimum}\footnote{\url{http://physionet.org/physiobank/database/mitdb/}} (24 instances), the Seating Arrangements Problem~\cite{martins2017lisbon} (21 instances), the Wireless Sensor Network Optimization Problem~\cite{kovasznai2018investigations, kovasznai2019portfolio} (18 instances).
    \item \textbf{MIPLIB}: All satisfiable 0-1 integer programs from the MIPLIB 2017 library and earlier MIPLIB releases\footnote{\url{https://zenodo.org/record/3870965}}, which contains 252 instances
    , provided in the literature~\cite{DBLP:conf/sat/0003BJ22}.
    \item \textbf{PB16}: The OPT-SMALL-INT benchmark from the most recent Pseudo-Boolean Competition 2016\footnote{\url{http://www.cril.univ-artois.fr/PB16/bench/PB16-used.tar}\label{pb16}}. 
    We filter out the duplicated instances that appear in both MIPLIB and PB16, resulting in $1524$ instances in the final. 
    PB16 contains different problem categories.
    We select those representatives (containing more than 30 instances) categories for finer-grained experimental analysis.
\end{itemize}
\subsection{Candidate Methods to Compare}
In the sequential track, we compare~\lspbodym{} with 7 state-or-the-art sequential PBO solvers, 
including 3 local search-based solvers: \lspbo{}, \lspbodeci{} and \nupbo{}, 3 complete non-commercial solvers: \hybrid{}, \pboihs{}, and \scip{}
and the commercial solver \gurobi{} (both complete and heuristic versions). 

In the parallel track, we compare \parlspbo{} with the academic solver \fiberscip{}, and the parallel version of the commercial solver \gurobi{}.

\begin{itemize}
    \item \lspbo ~\cite{SAT21-Lei-LSPBO}: the state-of-the-art SLS algorithm for solving PBO\footnote{\url{https://lcs.ios.ac.cn/\~caisw/Resource/LS-PBO/}}. 
    \item \lspbodeci ~\cite{DBLP:journals/corr/abs-2301-12251}: a recent SLS algorithm based on LS-PBO\footnote{\url{https://github.com/jiangluyu1998/DeciLS-PBO} (commit number: 3dce881)}.
    \item \nupbo ~\cite{chu2023towards}: a recent SLS algorithm based on LS-PBO, which established the latest state-of-the-art local search based PBO solving\footnote{\url{https://github.com/filyouzicha/NuPBO} (commit number: 821d901)}.
    \item \hybrid ~\cite{AAAI21-Devriendt-HybridPBO}: a recent core-guided PBO solver building upon RoundingSAT~\cite{DBLP:conf/ijcai/ElffersN18}\footnote{\url{https://zenodo.org/record/4043124} (version 2)}.
    \item \pboihs ~\cite{DBLP:conf/sat/0003BJ22}: a recent IHS PBO solver building upon RoundingSAT\footnote{\url{https://bitbucket.org/coreo-group/pbo-ihs-solver} (version 1.1)}.
    \item \gurobi ~\cite{gurobi2021gurobi}: one of the most powerful commercial MIP solvers. We use both its complete and heuristic versions\footnote{\url{https://www.gurobi.com/solutions/gurobi-optimizer} (version 10.0.0)}.
    \item \scip ~\cite{bestuzheva2021scip}: one of the fastest non-commercial solvers for MIP (the latest version 8.0.1)\footnote{\url{https://www.scipopt.org/index.php\#download} (version 8.0.1)},
    \item \fiberscip ~\cite{DBLP:journals/informs/ShinanoHVW18}: a parallel non-commercial MIP solvers based on SCIP (the latest version 1.0.0)\footnote{\url{https://ug.zib.de/index.php\#download} (version 1.0.0)}.
\end{itemize}

We download the latest version of all candidate methods to compare from their published links.
In all experiments, we always use their default parameter settings.


\subsection{Experimental Settings}
\lspbodym{} and \parlspbo{} are implemented in C++, and compiled with g++ (version 9.2.0) using the option ’-O2’. All experiments are carried out on a cluster with two AMD EPYC 7763 CPUs @ 2.45Ghz of 128 physical cores and 1TB memory running the operating system Ubuntu 20.04 LTS (64bit). 

As with the previous research on PBO solvers \cite{SAT21-Lei-LSPBO,DBLP:journals/corr/abs-2301-12251}, we set
the time limit for each run as $300$ and $3600$ seconds.
For each sequential randomized solver, we run $10$ times for each instance with different seeds from $\{0,1,\dots,9\}$, 
and select the median of the 10 runs as the final result. 
Without making any additional claims, the number of CPU cores that can be used for parallel solvers is set as 32.

For parameter tuning, we employed Sequential Model-based Algorithm Configuration (SMAC)~\cite{lindauer2022smac3}, conducting the tuning on 300 instances randomly selected from all the benchmarks, with a time limit set to 300 seconds. The parameter values obtained after tuning are listed in Table \ref{tab:5_1_parameter}\footnote{In fact, our solver is not sensitive to the parameter configurations. For example, a simple configuration ($K$=100000, $R$=100000, $inc$=1.1, $poolsize$=10, $\beta$=0.1, $\epsilon$=0.15) leads to a performance close to the one in Table 1, with a gap of $avg_{sc^*}$ less than 1\%.}.

\begin{table}[!h]
\centering
\renewcommand\arraystretch{1.3}
\caption{The parameter settings of our solvers.}
\label{tab:5_1_parameter}
\begin{tabular}{cccccccc}
\hline
Parameter & $K$ & $R$ & $inc$ & $poolsize$& $p^*$ & $\beta$ & $\epsilon$\\ \hline
Value & 566024 & 86295 & 1.15 & 18 & 0.58 & 0.03 & 0.144\\
\hline
\end{tabular}
\end{table}



Referring to the MaxSAT competition and previous research on PBO, we use $2$ metrics to evaluate the performance of each solver:
\begin{itemize}
    \item $\#win$: the number of instances that a solver finds the best solution among all solutions output by tested solvers (i.e., the number of winning instances).
    
    \item $avg_{sc^*}$: 
    Since 2017, in the incomplete track of recent MaxSAT Evaluations, the performance of various solvers is measured by competition scores.
    For an instance and a solver given, the competition score $sc$ is defined as $(1 + cost_{best}) /(1 + cost_{s})$, where $cost_{best}$ represents the objective value of the best solution found among all solvers, $cost_{s}$ represents the objective value of the solution found by the given solver.
    However, in PBO problem, the objective value of a solution may be negative, leading to an incorrect calculation of $sc$. 
    To address this issue, we modify slightly the definition of competition score:

\begin{equation*}
    sc^* = \frac{1 + cost_{best} + \sum_{c_i<0}{|c_i|}}{1 + cost_s + \sum_{c_i<0}{|c_i|}}
\end{equation*}
    
Adding all negative objective coefficients ensures the competition score of each instance is normalized in $[0, 1]$. 
We use $avg_{sc^*}$ to denote the average competition score of a solver.
\end{itemize}

We do not use average time as a metric because our primary focus is on the quality of the solution. If the quality of the solutions found is different, then the comparison based on run time would be misleading.

\input{Tables/5_1_LSPBOvsLSPBO-d}
\subsection{Performance Evaluations}
\subsubsection{The Sequential Track}
We first compare \lspbodym{} with \lspbo{}, and the results are shown in Table \ref{tab:5_2_lspboD_SOTA}. \lspbodym{} significantly improves \lspbo{} in terms of both $\#win$ and $avg_{sc^*}$ on all the benchmarks.

Further, we evaluate \lspbodym{} with other PBO solvers, as well as integer programming solvers. The results (Table \ref{tab:5_1_lspbo_lspboD}) indicate that \nupbo{} performs best for the Real-world benchmark, while \gurobi{} is the best on MIPLIB and PB16 benchmarks. \lspbodym{} cannot rival these two solvers, yet it is better than other PBO solvers. We note that the emphasis of this work is to develop an effective parallel method for PBO solvers. We choose \lspbo{} as the baseline as it is the typical local search PBO solver (\nupbo{} is also developed on top of it). We simply remedy its drawback to obtain \lspbodym{}, and do not perform other modifications. \nupbo{} was published very recently, and we believe our parallel method can be applied to \nupbo{} as well.




\subsubsection{The Parallel Track}
The comparative results of our parallel solver~\parlspbo{} with other parallel solvers are shown in Table \ref{tab:parallel-sota} (We only show $\#win$ due to the space limit). 
\parlspbo{} gives the best performance on all categories of the Real-World benchmark, and 3 categories of the PB16 benchmark, including Kexu, Logic Synthesis, and Prime. 

In terms of the Total instances, \parlspbo{} outperforms the non-commercial solver \fiberscip{}, and is competitive with the commercial solver \gurobi{}.
Comparing Table~\ref{tab:5_1_lspbo_lspboD} and Table~\ref{tab:parallel-sota}, it can be found that the gap between \parlspbo{} and \gurobi{} (32 threads) is decreasing compared with the gap between \lspbodym{} and \gurobi{} (1 thread), which indicates the effectiveness of our solver in parallel solving.

We also observe that \parlspbo{} outperforms the best sequential PBO solver \nupbo{} on all benchmarks (44 vs. 32, 171 vs. 156, and 1238 vs. 1002). Although this comparison is unfair (and thus we do not report it in the table), it indicates that by parallelization, the performance of PBO solvers can be significantly improved.

\input{Tables/5_3_parallel_sota}

\subsection{Effectiveness Analysis}
This subsection evaluates the effectiveness of the key strategies of \parlspbo{}. In Table \ref{tab:component}, we compare \parlspbo{} with its 2 variants:
\begin{itemize}
    \item $V_1$: to analyze the effectiveness of the solution-pool-based sharing, we modify \parlspbo{} by disabling the sharing mechanism and making each thread solve separately.
    \item $V_2$: to analyze the effectiveness of the global score mechanism, we modify \parlspbo{} by disabling the global score mechanism and using $score^*(x)$ directly in the local search.
\end{itemize}
As shown in Table \ref{tab:component}, \parlspbo{} outperforms other variations, confirming the effectiveness of the strategies.

\input{Tables/5_4_component}

\begin{figure}[htb!]
    \centering
    \begin{adjustbox}{minipage=\linewidth,scale=0.9}
    \begin{subfigure}[t]{0.49\linewidth}
    \centering
      \includegraphics[width=1\linewidth]{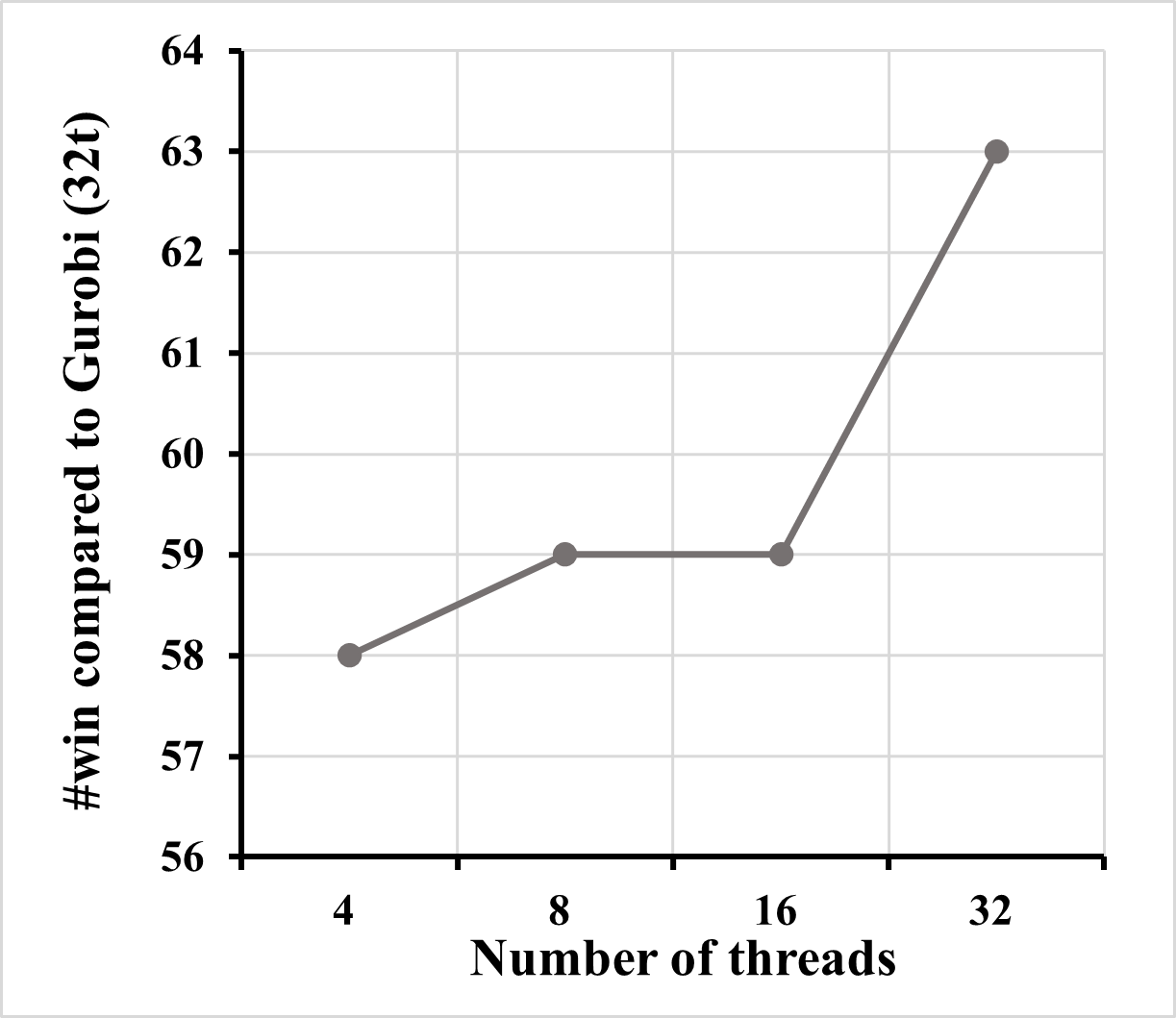}
      \caption{Real-World}
    \end{subfigure}
    \begin{subfigure}[t]{0.49\linewidth}
    \centering
      \includegraphics[width=1\linewidth]{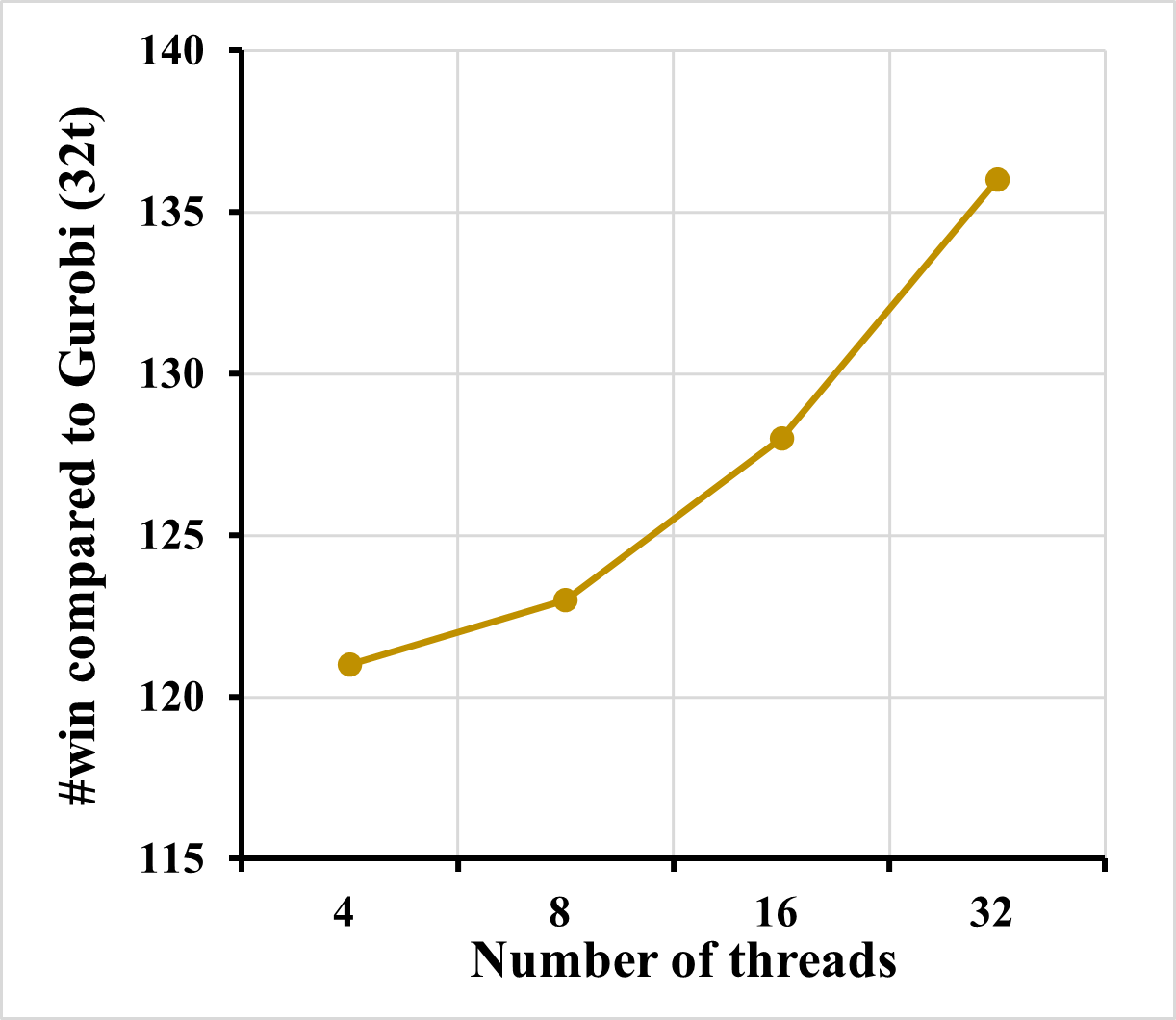}
      \caption{MIPLIB}
    \end{subfigure}
    \begin{subfigure}[t]{0.49\linewidth}
    \centering
      \includegraphics[width=1\linewidth]{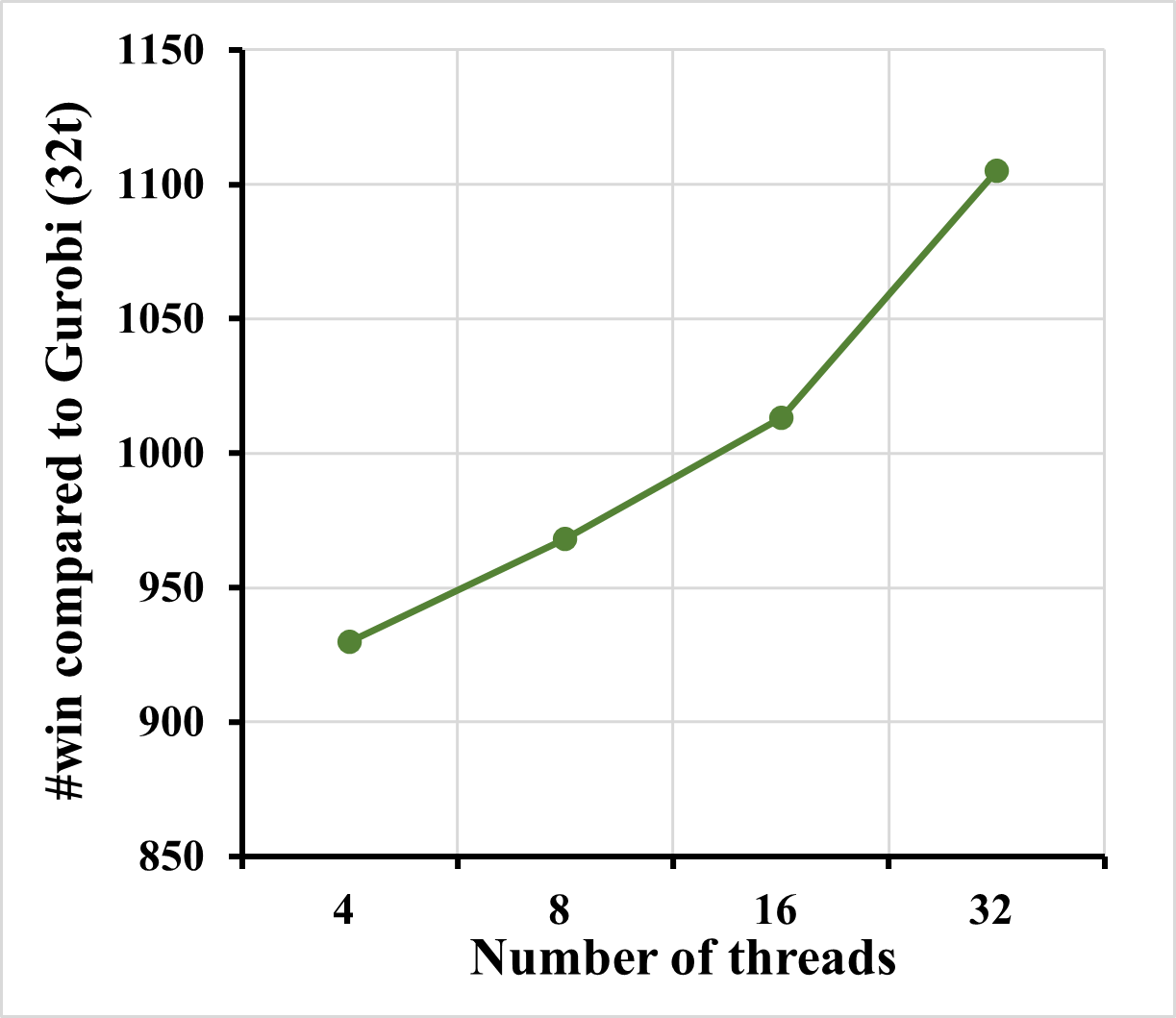}
      \caption{PB16}
    \end{subfigure}
    \begin{subfigure}[t]{0.49\linewidth}
    \centering
      \includegraphics[width=1\linewidth]{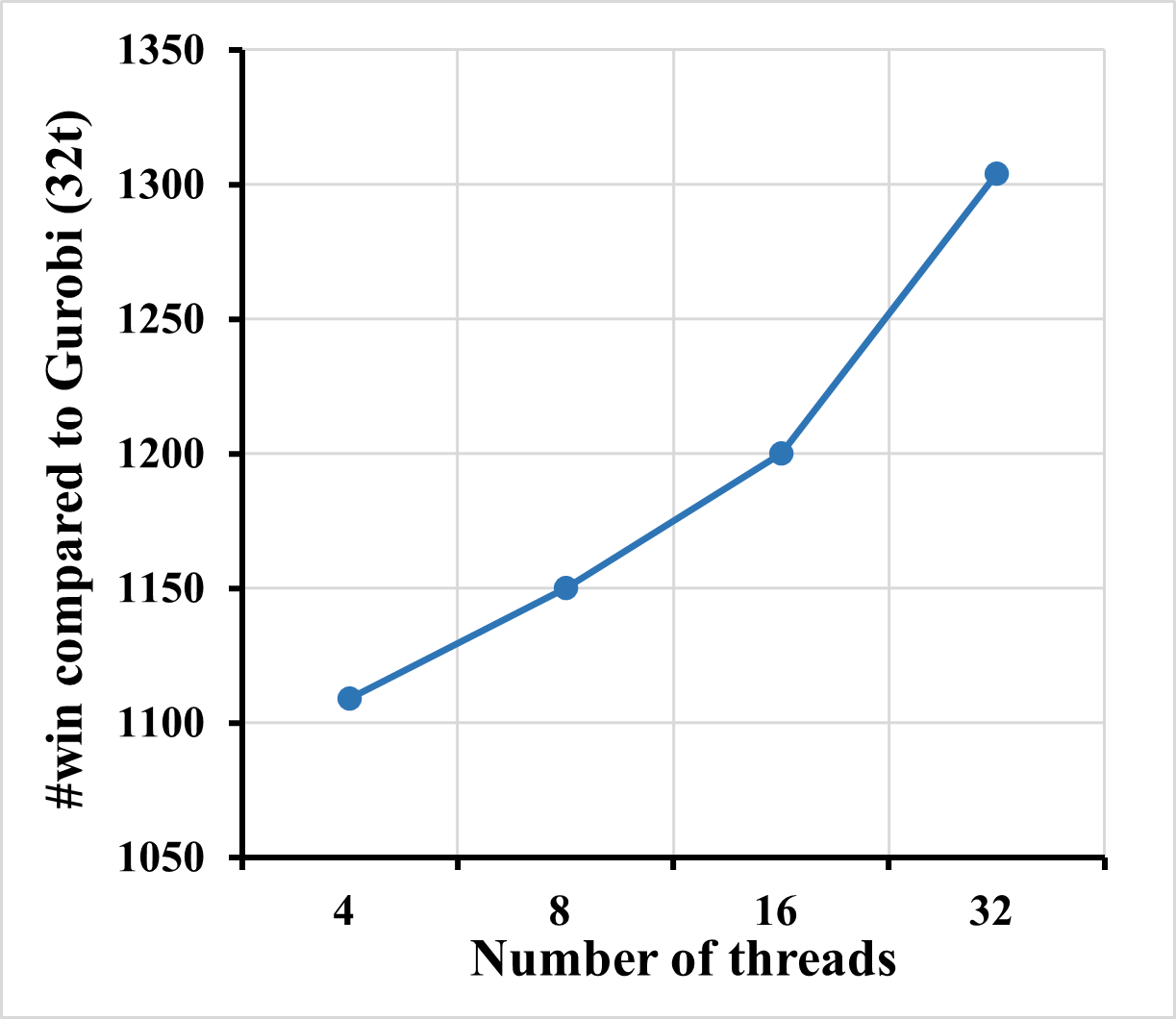}
      \caption{Total}
    \end{subfigure}
    \end{adjustbox}
    
    \caption{Scalability Analysis (Time limit is set to 300)}
    \label{fig_compare_lia}
\end{figure}

\subsection{Scalability Analysis}
\input{Tables/5_5_speed}
In order to analyze the scalability of \parlspbo{}, we choose \gurobi{} (complete version) with 32 threads as the comparison baseline to test the performance gap between different threads of \parlspbo{}. We report $\#win$ for threads set to $\{4, 8,16, 32\}$ compared to baseline. As is shown in Figure  \ref{fig_compare_lia}, in each benchmark, $\#win$ is gradually increasing, which verifies the scalability of \parlspbo{}.

%% file: Tables/5_1_LSPBOvsLSPBO-d.tex
\begin{table}[!t]
\centering
\renewcommand\arraystretch{1.1}
\setlength{\tabcolsep}{12pt}
\begin{threeparttable}
\caption{Evaluation between~\lspbodym{} and ~\lspbo{}. }
\label{tab:5_2_lspboD_SOTA}
\begin{tabular}{cccccc}
\hline
\multirow{2}{*}{Benchmark} & \multirow{2}{*}{\#Ins} & \multicolumn{2}{c}{\lspbo} & \multicolumn{2}{c}{\lspbodym} \\
\cmidrule(lr){3-4}\cmidrule(lr){5-6}
 &  & $\#win$ & $avg_{sc^*}$ & $\#win$ & $avg_{sc^*}$ \\
\hline
\multicolumn{6}{c}{cutoff=300s} \\
\hline
Real-World & 63 & 25 & 0.976 & \textbf{48} & \textbf{0.996} \\
miplib & 252 & 118 & 0.777 & \textbf{182} & \textbf{0.836} \\
PB16 & 1524 & 711 & 0.692 & \textbf{1124} & \textbf{0.776} \\
Total & 1839 & 854 & 0.713 & \textbf{1354} & \textbf{0.792} \\
\hline
\multicolumn{6}{c}{cutoff=3600s} \\
\hline
Real-World & 63 & \textbf{38} & \textbf{0.991} & 35 & 0.988 \\
miplib & 252 & 121 & 0.81 & \textbf{186} & \textbf{0.863} \\
PB16 & 1524 & 829 & 0.753 & \textbf{1189} & \textbf{0.825} \\
Total & 1839 & 988 & 0.769 & \textbf{1410} & \textbf{0.836} \\
\hline
\end{tabular}
\end{threeparttable}

\vspace{2em}

\small
\centering
\caption{\label{tab:5_1_lspbo_lspboD}
Performance evaluation between~\lspbodym{} and sequential SOTA solvers (The results of \lspbodeci{} are not presented due to the space limit. In fact, \lspbodeci{} is dominated by \nupbo{} and \lspbodym{}).}
\setlength\tabcolsep{2pt}
\renewcommand\arraystretch{1.2}
\begin{tabular}{cc|c|c|c|c|c|c|c}
\hline
\multirow{3}{*}{Benchmark} &\multirow{3}{*}{\#Ins} &\scip{} &\hybrid{} &\pboihs{} &\nupbo{} &\gurobi{}(Comp.) &\gurobi{}(Heur.) &\lspbodym{} \\
\cmidrule(lr){3-3}\cmidrule(lr){4-4}\cmidrule(lr){5-5}\cmidrule(lr){6-6}\cmidrule(lr){7-7}\cmidrule(lr){8-8}\cmidrule(lr){9-9}
& &$\#win$ &$\#win$ &$\#win$ &$\#win$ &$\#win$ &$\#win$ &$\#win$ \\
& &$avg_{sc^*}$ &$avg_{sc^*}$ &$avg_{sc^*}$ &$avg_{sc^*}$ &$avg_{sc^*}$ &$avg_{sc^*}$ &$avg_{sc^*}$ \\
\hline
\multicolumn{9}{c}{cutoff=300s}\\
\hline 
\multirow{2}{*}{Real-World} &\multirow{2}{*}{63} & 0 & 3 & 2 & \textbf{46} & 4 & 4 & 29 \\
 & & 0.126  & 0.109  & 0.266  & \textbf{0.972}  & 0.289  & 0.292  & 0.977 \\
\hline
\multirow{2}{*}{MIPLIB} &\multirow{2}{*}{252}& 88 & 53& 81& 116& 152& \textbf{165}& 101\\
 &   & 0.614  & 0.572  & 0.741  & 0.854  & 0.838  &\textbf{ 0.849}  & 0.803 \\
\hline
\multirow{2}{*}{PB16} &\multirow{2}{*}{1524} & 810 & 663 & 882& 980  & 1071 & \textbf{1072} & 842\\
&   & 0.687  & 0.624  & 0.804  & 0.813 & 0.84  & \textbf{0.84}  & 0.741 \\
\hline
\multicolumn{9}{c}{cutoff=3600s} \\
\hline
\multirow{2}{*}{Real-World} &\multirow{2}{*}{63}& 0& 11 & 5& \textbf{43}& 11& 9& 27\\
 &   & 0.171 & 0.494  & 0.401  & \textbf{0.997}  & 0.34  & 0.38  & 0.974 \\
 \hline
\multirow{2}{*}{MIPLIB} &\multirow{2}{*}{252}& 113& 65& 96& 113& 171 & \textbf{181}& 106\\
&   & 0.675  & 0.697  & 0.789  & 0.859  & 0.893 & \textbf{0.895}  & 0.831 \\
\hline
\multirow{2}{*}{PB16} &\multirow{2}{*}{1524}& 906& 729& 939& 1012& 1138& \textbf{1144}& 940\\
&   & 0.734  & 0.715  & 0.814  & 0.822  & 0.86  & \textbf{0.862}  & 0.797 \\
\hline
\end{tabular}
\end{table}

%% file: Tables/5_3_parallel_sota.tex
\begin{table}[!t]
\centering
\renewcommand\arraystretch{1.1}
\setlength{\tabcolsep}{10pt}
\begin{threeparttable}
\caption{Performance evaluation between~\parLSPBO{} and parallel SOTA solvers.}
		\label{tab:parallel-sota}
\begin{tabular}{ccc|cccc}
\hline
\multirow{2}{*}{Benchmark} &\multirow{2}{*}{Category} &
\multirow{2}{*}{\#ins}&\fiberscip{} &\multicolumn{2}{c}{\gurobi} &\parLSPBO{} \\  
\cline{5-6}
 & & & &Comp. &Heur. & \\ 
\hline
\multicolumn{7}{c}{cutoff=300s} \\
\hline
\multirow{3}{*}{Real-World} 
&MWCB &24 &0 &0  &0 &\textbf{24} \\
&WSNO &18 &0 &4  &4 &\textbf{18} \\
&SAP &21  &0 &0  &0 &\textbf{21} \\
&Total &63 &0 &4  &4 &\textbf{63} \\\hline
MIPLIB  &Total &252  &113  &\textbf{190} &180 &129\\\hline
\multirow{8}{*}{PB16} 
&Factor           &192    &\textbf{186} &\textbf{186} &\textbf{186} &172           \\
&Kexu             &40     &6            &10           &7            &\textbf{40}   \\
&Logic synthesis  &74     &71           &\textbf{73}  &72           &\textbf{73}   \\
&Market split     &40     &12           &\textbf{21}  &13           &5             \\
&Mps              &35     &30           &33           &\textbf{34}  &23            \\
&Numerical        &34     &13           &18           &\textbf{21}  &8             \\
&Prime            &156    &123          &128          &129          &\textbf{131}  \\
&Reduced mps      &273    &76           &145          &\textbf{150} &39            \\
&Total            &1524   &898          &\textbf{1147}&1143         &1100         \\ 
\hline
\multicolumn{7}{c}{cutoff=3600s} \\
\hline
\multirow{3}{*}{Real-World} 
&MWCB &24 &0 &5  &2 &\textbf{20}   \\
&WSNO &18 &0 &10 &10 &\textbf{18} \\
&SAP &21  &0 &0  &0 &\textbf{21}   \\
&Total &63&0 &15 &12 &\textbf{59} \\
\hline
MIPLIB  &Total &252 &129  &184 &\textbf{193} &140 \\
\hline
\multirow{8}{*}{PB16} 
&Factor           &192  &\textbf{186}  &\textbf{186}  &\textbf{186}  &182   \\
&Kexu             &40   &14  &17  &14  &\textbf{40}    \\
&Logic synthesis  &74   &72  &72  &72  &\textbf{74}    \\
&Market split     &40   &16  &\textbf{22} &12  &8     \\
&Mps              &35   &30  &\textbf{33} &\textbf{33} &25    \\
&Numerical        &34   &13  &19  &\textbf{25} &8     \\
&Prime            &156  &127 &130 &131 &\textbf{132}   \\
&Reduced mps      &273  &100 &150 &\textbf{160}  &43    \\
&Total            &1524 &995 &1198  &\textbf{1201} &1107            \\ 
\hline
\end{tabular}
\end{threeparttable}
\end{table}

%% file: Tables/5_4_component.tex
\begin{table}[!t]
\renewcommand\arraystretch{1.1}
\centering
\begin{threeparttable}
\caption{Performance evaluation between~\parLSPBO{} and its variants.}
\label{tab:component}
\begin{tabular}{cc|cccc|cccc}
\hline
\multirow{3}{*}{Benchmark} & \multirow{3}{*}{\#Ins} & \multicolumn{4}{c|}{$V_1$ vs. \parLSPBO{}} & \multicolumn{4}{c}{$V_2$ vs. \parLSPBO{}  } \\
 \cmidrule(lr){3-6}\cmidrule(lr){7-10}
 & & \multicolumn{2}{c}{$V_1$} &  \multicolumn{2}{c|}{\parLSPBO{}} & \multicolumn{2}{c}{$V_2$} &  \multicolumn{2}{c}{\parLSPBO{}}   \\
 \cmidrule(lr){3-4}\cmidrule(lr){5-6}\cmidrule(lr){7-8}\cmidrule(lr){9-10}
 & & \multicolumn{1}{c}{$\#win$} & \multicolumn{1}{c}{$avg_{sc^*}$} & \multicolumn{1}{c}{$\#win$} & \multicolumn{1}{c|}{$avg_{sc^*}$} & \multicolumn{1}{c}{$\#win$} & \multicolumn{1}{c}{$avg_{sc^*}$} & \multicolumn{1}{c}{$\#win$} & \multicolumn{1}{c}{$avg_{sc^*}$} \\
\hline
\multicolumn{10}{c}{cutoff=300s}\\
\hline
Real-World & 63 & 28 & 0.993 & \textbf{52} & \textbf{0.998} & 36 & 0.996 & \textbf{44} & \textbf{0.998} \\
MIPLIB & 252 & 137 & 0.857 & \textbf{199} & \textbf{0.868} & 166 & \textbf{0.871} & \textbf{185} & 0.870 \\
PB16 & 1524 & 1095 & 0.835 & \textbf{1231} & \textbf{0.847} & 1168 & 0.841 & \textbf{1182} & \textbf{0.844} \\
\hline
\multicolumn{10}{c}{cutoff=3600s}\\
\hline
Real-World & 63 & 19 & 0.981 & \textbf{62} & \textbf{1.0}  & 37 & 0.995 & \textbf{48} & \textbf{0.999} \\
MIPLIB & 252 & 145 & 0.870 & \textbf{203} & \textbf{0.897} & \textbf{181} &\textbf{0.897} & 180 & 0.893 \\
PB16 & 1524 & 1111 & 0.840 & \textbf{1251} & \textbf{0.854} & 1191 & 0.852 & \textbf{1214} & \textbf{0.852} \\
\hline

\end{tabular}
\end{threeparttable}
\end{table}

%% file: Tables/5_5_speed.tex

%% file: 6_Conclusion.tex
\section{Conclusions}

We proposed two local search solvers for the PBO problem: \lspbodym{} and \parlspbo{}. \lspbodym{} is an enhanced version of the \lspbo{} solver, incorporating a dynamic scoring mechanism. \parlspbo{} is a parallel solver with a solution pool collecting good solutions from multiple threads. The solution pool guides the local search process by providing better starting points and utilizing polarity information from high-quality solutions to improve the scoring function. Experimental results show that our parallel solver has significantly better performance than all sequential solvers and exhibits strong competitiveness against the parallel versions of \gurobi{}.

The ideas of this work can be applied to other problems, particularly including SAT and MaxSAT. It is also interesting to implement a distributed version of \parlspbo{} for cloud computation.

%% file: lipics-v2021-sample-article.bbl
\begin{thebibliography}{10}

\bibitem{DBLP:conf/sat/BalintS12}
Adrian Balint and Uwe Sch{\"{o}}ning.
\newblock Choosing probability distributions for stochastic local search and the role of make versus break.
\newblock In Alessandro Cimatti and Roberto Sebastiani, editors, {\em Theory and Applications of Satisfiability Testing - {SAT} 2012 - 15th International Conference, Trento, Italy, June 17-20, 2012. Proceedings}, volume 7317 of {\em Lecture Notes in Computer Science}, pages 16--29. Springer, 2012.
\newblock \href {https://doi.org/10.1007/978-3-642-31612-8\_3} {\path{doi:10.1007/978-3-642-31612-8\_3}}.

\bibitem{TR95-Barth-DPLinearPBO}
Peter Barth.
\newblock A davis-putnam based enumeration algorithm for linear pseudo-boolean optimization.
\newblock Technical report, Max Plank Institute for Computer Science, 1995.

\bibitem{STACS94-Benhamou-ProofCardinality}
Belaid Benhamou, Lakhdar Sais, and Pierre Siegel.
\newblock Two proof procedures for a cardinality based language in propositional calculus.
\newblock In Patrice Enjalbert, Ernst~W. Mayr, and Klaus~W. Wagner, editors, {\em {STACS} 94, 11th Annual Symposium on Theoretical Aspects of Computer Science, Caen, France, February 24-26, 1994, Proceedings}, volume 775 of {\em Lecture Notes in Computer Science}, pages 71--82. Springer, 1994.
\newblock \href {https://doi.org/10.1007/3-540-57785-8\_132} {\path{doi:10.1007/3-540-57785-8\_132}}.

\bibitem{berg2017minimum}
Jeremias Berg, Emilia Oikarinen, Matti J{\"a}rvisalo, and Kai Puolam{\"a}ki.
\newblock Minimum-width confidence bands via constraint optimization.
\newblock In {\em International Conference on Principles and Practice of Constraint Programming}, pages 443--459. Springer, 2017.

\bibitem{bestuzheva2021scip}
Ksenia Bestuzheva, Mathieu Besan{\c{c}}on, Wei-Kun Chen, Antonia Chmiela, Tim Donkiewicz, Jasper van Doornmalen, Leon Eifler, Oliver Gaul, Gerald Gamrath, Ambros Gleixner, et~al.
\newblock The scip optimization suite 8.0.
\newblock {\em arXiv preprint arXiv:2112.08872}, 2021.

\bibitem{DBLP:conf/sat/CaiLS15}
Shaowei Cai, Chuan Luo, and Kaile Su.
\newblock Ccanr: {A} configuration checking based local search solver for non-random satisfiability.
\newblock In Marijn Heule and Sean~A. Weaver, editors, {\em Theory and Applications of Satisfiability Testing - {SAT} 2015 - 18th International Conference, Austin, TX, USA, September 24-27, 2015, Proceedings}, volume 9340 of {\em Lecture Notes in Computer Science}, pages 1--8. Springer, 2015.
\newblock \href {https://doi.org/10.1007/978-3-319-24318-4\_1} {\path{doi:10.1007/978-3-319-24318-4\_1}}.

\bibitem{Chen-2016-rank}
Yuning Chen and Jin{-}Kao Hao.
\newblock Memetic search for the generalized quadratic multiple knapsack problem.
\newblock {\em {IEEE} Trans. Evol. Comput.}, 20(6):908--923, 2016.
\newblock \href {https://doi.org/10.1109/TEVC.2016.2546340} {\path{doi:10.1109/TEVC.2016.2546340}}.

\bibitem{chen2023prs}
Zhihan Chen, Xindi Zhang, Yuhang Qian, and Shaowei Cai.
\newblock Prs: A new parallel/distributed framework for sat.
\newblock {\em SAT COMPETITION 2023}, page~39, 2023.

\bibitem{DBLP:conf/aaai/Chu0L23}
Yi~Chu, Shaowei Cai, and Chuan Luo.
\newblock Nuwls: Improving local search for (weighted) partial maxsat by new weighting techniques.
\newblock In Brian Williams, Yiling Chen, and Jennifer Neville, editors, {\em Thirty-Seventh {AAAI} Conference on Artificial Intelligence, {AAAI} 2023, Thirty-Fifth Conference on Innovative Applications of Artificial Intelligence, {IAAI} 2023, Thirteenth Symposium on Educational Advances in Artificial Intelligence, {EAAI} 2023, Washington, DC, USA, February 7-14, 2023}, pages 3915--3923. {AAAI} Press, 2023.
\newblock URL: \url{https://ojs.aaai.org/index.php/AAAI/article/view/25505}.

\bibitem{chu2023towards}
Yi~Chu, Shaowei Cai, Chuan Luo, Zhendong Lei, and Cong Peng.
\newblock Towards more efficient local search for pseudo-boolean optimization.
\newblock In {\em 29th International Conference on Principles and Practice of Constraint Programming (CP 2023)}. Schloss-Dagstuhl-Leibniz Zentrum f{\"u}r Informatik, 2023.

\bibitem{DAC95-Coudert-BnBPBOMHS}
Olivier Coudert and Jean~Christophe Madre.
\newblock New ideas for solving covering problems.
\newblock In Bryan Preas, editor, {\em Proceedings of the 32st Conference on Design Automation, San Francisco, California, USA, Moscone Center, June 12-16, 1995}, pages 641--646. {ACM} Press, 1995.
\newblock \href {https://doi.org/10.1145/217474.217603} {\path{doi:10.1145/217474.217603}}.

\bibitem{AAAI21-Devriendt-HybridPBO}
Jo~Devriendt, Stephan Gocht, Emir Demirovic, Jakob Nordstr{\"{o}}m, and Peter~J. Stuckey.
\newblock Cutting to the core of pseudo-boolean optimization: Combining core-guided search with cutting planes reasoning.
\newblock In {\em Thirty-Fifth {AAAI} Conference on Artificial Intelligence, {AAAI} 2021, Thirty-Third Conference on Innovative Applications of Artificial Intelligence, {IAAI} 2021, The Eleventh Symposium on Educational Advances in Artificial Intelligence, {EAAI} 2021, Virtual Event, February 2-9, 2021}, pages 3750--3758. {AAAI} Press, 2021.
\newblock URL: \url{https://ojs.aaai.org/index.php/AAAI/article/view/16492}.

\bibitem{een2006translating}
Niklas E{\'e}n and Niklas S{\"o}rensson.
\newblock Translating pseudo-boolean constraints into sat.
\newblock {\em Journal on Satisfiability, Boolean Modeling and Computation}, 2(1-4):1--26, 2006.

\bibitem{IJCAI18-Elffers-RoundingSAT}
Jan Elffers and Jakob Nordstr{\"{o}}m.
\newblock Divide and conquer: Towards faster pseudo-boolean solving.
\newblock In J{\'{e}}r{\^{o}}me Lang, editor, {\em Proceedings of the Twenty-Seventh International Joint Conference on Artificial Intelligence, {IJCAI} 2018, July 13-19, 2018, Stockholm, Sweden}, pages 1291--1299. ijcai.org, 2018.
\newblock \href {https://doi.org/10.24963/ijcai.2018/180} {\path{doi:10.24963/ijcai.2018/180}}.

\bibitem{DBLP:conf/ijcai/ElffersN18}
Jan Elffers and Jakob Nordstr{\"{o}}m.
\newblock Divide and conquer: Towards faster pseudo-boolean solving.
\newblock In J{\'{e}}r{\^{o}}me Lang, editor, {\em Proceedings of the Twenty-Seventh International Joint Conference on Artificial Intelligence, {IJCAI} 2018, July 13-19, 2018, Stockholm, Sweden}, pages 1291--1299. ijcai.org, 2018.
\newblock \href {https://doi.org/10.24963/ijcai.2018/180} {\path{doi:10.24963/ijcai.2018/180}}.

\bibitem{fleury2020cadical}
Armin Biere Katalin Fazekas~Mathias Fleury and Maximilian Heisinger.
\newblock Cadical, kissat, paracooba, plingeling and treengeling entering the sat competition 2020.
\newblock {\em SAT COMPETITION}, 2020:50, 2020.

\bibitem{DBLP:conf/sat/FriouxBSK17}
Ludovic~Le Frioux, Souheib Baarir, Julien Sopena, and Fabrice Kordon.
\newblock Painless: {A} framework for parallel {SAT} solving.
\newblock In Serge Gaspers and Toby Walsh, editors, {\em Theory and Applications of Satisfiability Testing - {SAT} 2017 - 20th International Conference, Melbourne, VIC, Australia, August 28 - September 1, 2017, Proceedings}, volume 10491 of {\em Lecture Notes in Computer Science}, pages 233--250. Springer, 2017.
\newblock \href {https://doi.org/10.1007/978-3-319-66263-3\_15} {\path{doi:10.1007/978-3-319-66263-3\_15}}.

\bibitem{gurobi2021gurobi}
LLC Gurobi~Optimization.
\newblock Gurobi optimizer reference manual, 2021.

\bibitem{DBLP:conf/ecai/IserBJ23}
Markus Iser, Jeremias Berg, and Matti J{\"{a}}rvisalo.
\newblock Oracle-based local search for pseudo-boolean optimization.
\newblock In Kobi Gal, Ann Now{\'{e}}, Grzegorz~J. Nalepa, Roy Fairstein, and Roxana Radulescu, editors, {\em {ECAI} 2023 - 26th European Conference on Artificial Intelligence, September 30 - October 4, 2023, Krak{\'{o}}w, Poland - Including 12th Conference on Prestigious Applications of Intelligent Systems {(PAIS} 2023)}, volume 372 of {\em Frontiers in Artificial Intelligence and Applications}, pages 1124--1131. {IOS} Press, 2023.
\newblock \href {https://doi.org/10.3233/FAIA230387} {\path{doi:10.3233/FAIA230387}}.

\bibitem{DBLP:journals/corr/abs-2301-12251}
Luyu Jiang, Dantong Ouyang, Qi~Zhang, and Liming Zhang.
\newblock Decils-pbo: an effective local search method for pseudo-boolean optimization.
\newblock {\em CoRR}, abs/2301.12251, 2023.
\newblock \href {https://arxiv.org/abs/2301.12251} {\path{arXiv:2301.12251}}, \href {https://doi.org/10.48550/arXiv.2301.12251} {\path{doi:10.48550/arXiv.2301.12251}}.

\bibitem{kovasznai2018investigations}
Gergely Kov{\'a}sznai, Bal{\'a}zs Erd{\'e}lyi, and Csaba Bir{\'o}.
\newblock Investigations of graph properties in terms of wireless sensor network optimization.
\newblock In {\em 2018 IEEE International Conference on Future IoT Technologies (Future IoT)}, pages 1--8. IEEE, 2018.

\bibitem{kovasznai2019portfolio}
Gergely Kov{\'a}sznai, Kriszti{\'a}n Gajd{\'a}r, and Laura Kov{\'a}cs.
\newblock Portfolio sat and smt solving of cardinality constraints in sensor network optimization.
\newblock In {\em 2019 21st International Symposium on Symbolic and Numeric Algorithms for Scientific Computing (SYNASC)}, pages 85--91. IEEE, 2019.

\bibitem{le2010sat4j}
Daniel Le~Berre and Anne Parrain.
\newblock The sat4j library, release 2.2.
\newblock {\em Journal on Satisfiability, Boolean Modeling and Computation}, 7(2-3):59--64, 2010.

\bibitem{DBLP:conf/ijcai/LeiC18}
Zhendong Lei and Shaowei Cai.
\newblock Solving (weighted) partial maxsat by dynamic local search for {SAT}.
\newblock In J{\'{e}}r{\^{o}}me Lang, editor, {\em Proceedings of the Twenty-Seventh International Joint Conference on Artificial Intelligence, {IJCAI} 2018, July 13-19, 2018, Stockholm, Sweden}, pages 1346--1352. ijcai.org, 2018.
\newblock \href {https://doi.org/10.24963/ijcai.2018/187} {\path{doi:10.24963/ijcai.2018/187}}.

\bibitem{SAT21-Lei-LSPBO}
Zhendong Lei, Shaowei Cai, Chuan Luo, and Holger~H. Hoos.
\newblock Efficient local search for pseudo boolean optimization.
\newblock In Chu{-}Min Li and Felip Many{\`{a}}, editors, {\em Theory and Applications of Satisfiability Testing - {SAT} 2021 - 24th International Conference, Barcelona, Spain, July 5-9, 2021, Proceedings}, volume 12831 of {\em Lecture Notes in Computer Science}, pages 332--348. Springer, 2021.
\newblock \href {https://doi.org/10.1007/978-3-030-80223-3\_23} {\path{doi:10.1007/978-3-030-80223-3\_23}}.

\bibitem{DBLP:conf/sat/LiL12}
Chu~Min Li and Yu~Li.
\newblock Satisfying versus falsifying in local search for satisfiability - (poster presentation).
\newblock In Alessandro Cimatti and Roberto Sebastiani, editors, {\em Theory and Applications of Satisfiability Testing - {SAT} 2012 - 15th International Conference, Trento, Italy, June 17-20, 2012. Proceedings}, volume 7317 of {\em Lecture Notes in Computer Science}, pages 477--478. Springer, 2012.
\newblock \href {https://doi.org/10.1007/978-3-642-31612-8\_43} {\path{doi:10.1007/978-3-642-31612-8\_43}}.

\bibitem{DAC97-Liao-BnBPBOLPR}
Stan~Y. Liao and Srinivas Devadas.
\newblock Solving covering problems using lpr-based lower bounds.
\newblock In Ellen~J. Yoffa, Giovanni~De Micheli, and Jan~M. Rabaey, editors, {\em Proceedings of the 34st Conference on Design Automation, Anaheim, California, USA, Anaheim Convention Center, June 9-13, 1997}, pages 117--120. {ACM} Press, 1997.
\newblock \href {https://doi.org/10.1145/266021.266046} {\path{doi:10.1145/266021.266046}}.

\bibitem{lindauer2022smac3}
Marius Lindauer, Katharina Eggensperger, Matthias Feurer, Andr{\'e} Biedenkapp, Difan Deng, Carolin Benjamins, Tim Ruhkopf, Ren{\'e} Sass, and Frank Hutter.
\newblock Smac3: A versatile bayesian optimization package for hyperparameter optimization.
\newblock {\em Journal of Machine Learning Research}, 23(54):1--9, 2022.

\bibitem{marques2021conflict}
Joao Marques-Silva, In{\^e}s Lynce, and Sharad Malik.
\newblock Conflict-driven clause learning sat solvers.
\newblock In {\em Handbook of satisfiability}, pages 133--182. 2021.

\bibitem{martins2011exploiting}
Ruben Martins, Vasco Manquinho, and In{\^e}s Lynce.
\newblock Exploiting cardinality encodings in parallel maximum satisfiability.
\newblock In {\em 2011 IEEE 23rd International Conference on Tools with Artificial Intelligence}, pages 313--320. IEEE, 2011.

\bibitem{martins2012parallel}
Ruben Martins, Vasco Manquinho, and In{\^e}s Lynce.
\newblock Parallel search for maximum satisfiability.
\newblock {\em AI Communications}, 25(2):75--95, 2012.

\bibitem{SAT14-Martins-OpenWBO}
Ruben Martins, Vasco~M. Manquinho, and In{\^{e}}s Lynce.
\newblock Open-wbo: {A} modular maxsat solver,.
\newblock In Carsten Sinz and Uwe Egly, editors, {\em Theory and Applications of Satisfiability Testing - {SAT} 2014 - 17th International Conference, Held as Part of the Vienna Summer of Logic, {VSL} 2014, Vienna, Austria, July 14-17, 2014. Proceedings}, volume 8561 of {\em Lecture Notes in Computer Science}, pages 438--445. Springer, 2014.
\newblock \href {https://doi.org/10.1007/978-3-319-09284-3\_33} {\path{doi:10.1007/978-3-319-09284-3\_33}}.

\bibitem{martins2017lisbon}
Ruben Martins and Justine Sherry.
\newblock Lisbon wedding: seating arrangements using maxsat.
\newblock {\em MaxSAT Evaluation}, pages 25--26, 2017.

\bibitem{Chapter21-Roussel-PBandCardinality}
Olivier Roussel and Vasco~M. Manquinho.
\newblock Pseudo-boolean and cardinality constraints.
\newblock In Armin Biere, Marijn Heule, Hans van Maaren, and Toby Walsh, editors, {\em Handbook of Satisfiability - Second Edition}, volume 336 of {\em Frontiers in Artificial Intelligence and Applications}, pages 1087--1129. {IOS} Press, 2021.
\newblock \href {https://doi.org/10.3233/FAIA201012} {\path{doi:10.3233/FAIA201012}}.

\bibitem{DBLP:journals/informs/ShinanoHVW18}
Yuji Shinano, Stefan Heinz, Stefan Vigerske, and Michael Winkler.
\newblock Fiberscip - {A} shared memory parallelization of {SCIP}.
\newblock {\em {INFORMS} J. Comput.}, 30(1):11--30, 2018.
\newblock \href {https://doi.org/10.1287/ijoc.2017.0762} {\path{doi:10.1287/ijoc.2017.0762}}.

\bibitem{DBLP:conf/sat/0003BJ22}
Pavel Smirnov, Jeremias Berg, and Matti J{\"{a}}rvisalo.
\newblock Improvements to the implicit hitting set approach to pseudo-boolean optimization.
\newblock In Kuldeep~S. Meel and Ofer Strichman, editors, {\em 25th International Conference on Theory and Applications of Satisfiability Testing, {SAT} 2022, August 2-5, 2022, Haifa, Israel}, volume 236 of {\em LIPIcs}, pages 13:1--13:18. Schloss Dagstuhl - Leibniz-Zentrum f{\"{u}}r Informatik, 2022.
\newblock \href {https://doi.org/10.4230/LIPIcs.SAT.2022.13} {\path{doi:10.4230/LIPIcs.SAT.2022.13}}.

\bibitem{tchinda2021hkis}
Rodrigue~Konan Tchinda and Cl{\'e}mentin~Tayou Djamegni.
\newblock Hkis, hcad, pakis and painless exmaplelcmdistchronobt in the sc21.
\newblock {\em SAT COMPETITION}, 2021:26, 2021.

\end{thebibliography}
